\documentclass[letterpaper, 10 pt, conference]{ieeeconf}

\IEEEoverridecommandlockouts                              

\usepackage[l2tabu,orthodox]{nag}

\usepackage[
backend=biber, 
bibencoding=utf8,
style=numeric-comp, 
sorting=none, 
natbib=true, 
doi=false, 
isbn=false, 
url=false, 
eprint=false,
maxnames=50,
maxcitenames=1, 
mincitenames=1,
firstinits=true,
]{biblatex}

\usepackage[draft,pdftex,colorlinks]{hyperref}

\usepackage[printonlyused]{acronym}

\usepackage{siunitx}
\usepackage[all]{nowidow}

\usepackage[dvipsnames]{xcolor}

\usepackage{lipsum}

\usepackage[pdftex]{graphicx}

\usepackage{epstopdf}

\usepackage{import}

\graphicspath{{./latexGoodPractices/}}

\usepackage{booktabs}

\usepackage{tabu}

\usepackage{tabularray} 

\usepackage{amssymb,amsfonts,amsmath,amscd}

\usepackage{bm} 

\newcommand{\bbm}{\begin{bmatrix}}
\newcommand{\ebm}{\end{bmatrix}}

\usepackage[normalem]{ulem} 

\makeatletter
\usepackage[utf8]{inputenc}
\usepackage{commath}
\usepackage{mathtools}
\usepackage{mathabx}
\usepackage{siunitx}
\usepackage[ruled,vlined]{algorithm2e}
\usepackage[lofdepth,lotdepth]{subfig}
\usepackage{tikz}
\usepackage{dblfloatfix}
\usepackage{xcolor,mdframed}
\usepackage{multirow}

\let\oldtheequation\theequation
\renewcommand\tagform@[1]{\maketag@@@{\ignorespaces#1\unskip\@@italiccorr}}
\renewcommand\theequation{(\oldtheequation)}
\makeatother

\DeclareSIUnit{\deg}{deg}
\DeclareSIUnit{\rad}{rad}
\DeclareSIUnit{\min}{min}
\DeclareSIUnit{\bps}{Bps}

\usepackage[belowskip=0pt,aboveskip=1pt]{caption} 
\graphicspath{{media/}}

\newcommand{\ie}{i.e., }
\newcommand{\eg}{e.g., }

\usepackage{amssymb}
\usepackage{pifont}

\acrodef{lidar}{Light Detection And Ranging}
\acrodefplural{lidar}{Light Detection And Ranging}
\acrodef{ICP}{Iterative Closest Point}
\acrodef{DOF}{Degrees Of Freedom}
\acrodef{RTS}{Robotic Total Station}
\acrodefplural{RTS}{Robotic Total Stations}
\acrodef{GNSS}{Global Navigation Satellite System}
\acrodef{RTK}{Real Time Kinematics}
\acrodef{GCP}{Ground Control Point}
\acrodef{UAV}{Unmanned Aerial Vehicle}
\acrodef{IQR}{Interquartile Range}
\acrodef{IMU}{Inertial Measurement Unit}
\acrodef{SLAM}{Simultaneous Localization and Mapping}
\acrodef{INS}{Inertial Navigation System}
\acrodef{UGV}{Uncrewed Ground Vehicle}
\acrodef{NTP}{Network Time Protocol}
\acrodef{HD}{High-Definition}
\acrodef{TLS}{Terrestrial Laser Scanner}
\acrodef{CAD}{Computer-Aided Design}
\acrodef{URDF}{Unified Robot Description Format}
\acrodef{RTS-GT}{Robotic Total Stations Ground Truthing dataset}

\newcommand{\dataset}{\ac{RTS-GT}}

\DeclareUnicodeCharacter{0308}{¨} 

\usepackage{rotating}
\usepackage{graphicx}
\usepackage{adjustbox}
\newcolumntype{R}[2]{%
    >{\adjustbox{angle=#1,lap=\width-(#2)}\bgroup}%
    l%
    <{\egroup}%
}

\usepackage{diagbox}

\usepackage{highlight}
\usepackage{colortbl}

\usepackage{tabularx}

\begin{document}
\title{\LARGE \textbf{RTS-GT: Robotic Total Stations Ground Truthing dataset}}

\author{Maxime Vaidis$^{1}$, Mohsen Hassanzadeh Shahraji, Effie Daum, William Dubois,\\Philippe Giguère, François Pomerleau$^{1}$
}

\linepenalty=3000
\addtolength{\textfloatsep}{-0.1in}

\thispagestyle{empty}
\pagestyle{empty}

\twocolumn[{%
\renewcommand\twocolumn[1][]{#1}
\maketitle
\vspace{-5mm}
\begin{center}
    \centering
    \captionsetup{type=figure}
    \includegraphics[width=\linewidth, trim={0 15 0 26}, clip]{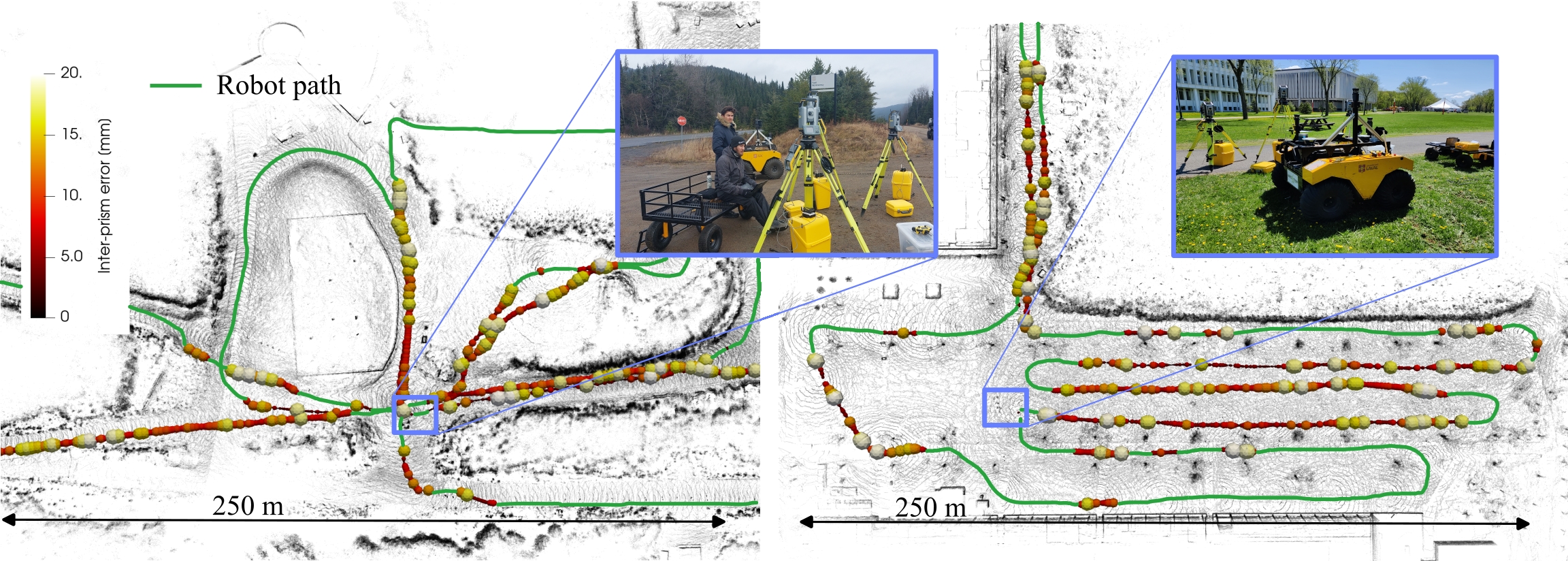}
    \captionof{figure}{Example of two areas provided in the dataset (bird's-eye view). 
    \emph{Left:} a forest. \emph{Right:} a park on a campus.
    Maps in black are based on lidar scans, while the colored spheres represent the scaled-up uncertainty of the provided ground truth in millimeters.  
    A Clearpath Warthog platform was used during these experiments, whose paths are illustrated in green.}
    \label{fig:intro}
\end{center}
}]
{
  \renewcommand{\thefootnote}%
    {\fnsymbol{footnote}}
  \footnotetext{$^{1}$Northern Robotics Laboratory, Université Laval, Québec City, Canada,
  {\texttt{\small{$\{$maxime.vaidis, francois.pomerleau$\}$ @norlab.ulaval.ca}}}}
  \footnotetext{*This research was supported by the  Natural  Sciences and Engineering  Research  Council of  Canada  (NSERC)  through the grant CRDPJ 527642-18 SNOW (Self-driving Navigation Optimized for Winter).}
}


\begin{abstract}

Numerous datasets and benchmarks exist to assess and compare \ac{SLAM} algorithms. 
Nevertheless, their precision must follow the rate at which \ac{SLAM} algorithms improved in recent years.
Moreover, current datasets fall short of comprehensive data-collection protocol for reproducibility and the evaluation of the precision or accuracy of the recorded trajectories.
With this objective in mind, we proposed the \dataset{} dataset to support localization research with the generation of six-\ac{DOF} ground truth trajectories.
This novel dataset includes six-\ac{DOF} ground truth trajectories generated using a system of three \acp{RTS} tracking moving robotic platforms.
Furthermore, we compare the performance of the \ac{RTS}-based system to a \ac{GNSS}-based setup. 
The dataset comprises around sixty experiments conducted in various conditions over a period of \num{17} months, and encompasses over \num{49} kilometers of trajectories, making it the most extensive dataset of \ac{RTS}-based measurements to date.
Additionally, we provide the precision of all poses for each experiment, a feature not found in the current state-of-the-art datasets.
Our results demonstrate that \acp{RTS} provide measurements that are \num{22} times more stable than \ac{GNSS} in various environmental settings, making them a valuable resource for \ac{SLAM} benchmark development.

\end{abstract}

\acresetall
\section{Introduction}

Accurate and precise ground truth trajectories in open-source datasets are essential to evaluate \ac{SLAM} algorithms~\cite{Li2022}. 
Motion capture such as Vicon or OptiTrack systems have become the \emph{de facto} standard to generate such ground truth in indoor environments~\cite{TUMVI2018}.
Nonetheless, they are not suitable to be deployed outside due to direct sunlight corrupting the readings and the need to cover significantly larger areas.
For outdoor deployments, the majority of datasets use \ac{GNSS} in \ac{RTK} mode integrated with \ac{INS} to compute the reference trajectories within several centimeters accuracy~\cite{KAIST2018, Oxford2020, NCLT2015}.
Meanwhile, such localization systems are vulnerable to \ac{GNSS} outages, as seen on the KITTI dataset~\cite{KITTI2012}.
A few datasets rely on \ac{HD} maps from a terrestrial laser scanner to obtain the reference within centimeter accuracy~\cite{Ebadi2022, Hilti2023}.
However, surveying remains an open challenge, especially in off-road terrains where reference planes are scarce, making it challenging to align scans.

In recent years, a limited number of datasets used a single \ac{RTS} to generate such references indoors or outdoors with various platforms~\cite{UZH2019,Hilti2022}. 
\ac{RTS} is a surveying instrument capable of measuring the position of a reflective target (hereafter \emph{prism}) with millimeter precision in two modes: 1) static when the prism is fixed, and 2) dynamic when the prism is in motion.
Moreover, two types of prism exist, active and passive.
An active prism is equipped with LEDs that emit a unique infrared signature.
This active signature enables multiple \acp{RTS} to automatically track different prisms within their field of view. 
Regardless of their accuracy and robustness, a single \ac{RTS} can only provide the reference 3D position of the tracked platform.
This limitation arises since only one prism can be tracked while the robot is in motion.
By using three passive prisms, it is possible to obtain the six-\ac{DOF} reference trajectory by collecting manually the data when the robot is static~\cite{Challenge2012}, which makes the data collection procedure cumbersome. 
To the best of our knowledge, active prisms were never used in any released datasets providing six-\ac{DOF} reference trajectories generated by \acp{RTS} in dynamic mode.
Moreover, new research done recently enables estimation of the uncertainty for a multi-\ac{RTS} setup~\cite{Vaidis2023Iros}.
This uncertainty on ground truth is not available in state-of-the-art \ac{SLAM} datasets, which can lead to unbiased comparisons as many \ac{SLAM} algorithms are approaching centimeter-level accuracy~\cite{Campos2020}.

The key motivation for this work is to provide a high-quality dataset that covers a variety of challenging environments and further motivates ground truth generation for \ac{SLAM} algorithms. 
As shown in our previous research~\cite{Vaidis2023Icra, Vaidis2023Iros}, generating a reliable six-\ac{DOF} reference trajectories of a moving robot with three \acp{RTS} is now feasible.
As a follow-up, we present \dataset{}, a dataset providing six-\ac{DOF} reference trajectories originating from two different types of setup, three \acp{RTS} and three \ac{GNSS} receivers.
The \dataset{} dataset was collected during \num{17} months in diverse weather conditions, totaling over \num{49} kilometers of trajectories.
With this dataset, we provide and compare the precision and the reproducibility of the setups during multiple weather conditions and in different environments as shown in~\autoref{fig:intro}, demonstrating that \acp{RTS} are more reliable than \ac{GNSS} to generate ground truth trajectories.

\section{Related work}
\label{sec:related_work}
\tabulinesep=0.5mm
\begin{table*}[!htbp]
        \scriptsize
	\centering
    \vspace{1mm}
	\caption{Comparison of public datasets containing ground truth trajectories. 
        The top half of the table shows the most popular datasets used for SLAM evaluation. 
        The bottom half shows public datasets that contain ground truth generated by the most accurate setups. 
        The \dataset{} dataset is the only that allows dynamic six-DOF ground truth generation with RTSs on a large scale. The symbol $^\ast$ indicates the value is estimated by us. The letter \emph{O} means \emph{Outside} and the letter \emph{I} means \emph{Inside}.}
	\label{tab:related_work}
	\begin{tabu}{X[6,l]X[4.8,l]X[2,l]
                    X[0.6,c]X[0.6,c]X[0.6,c]X[0.6,c]X[0.6,c]
                    X[3.5,c]X[0.5,c]X[2.8,c]X[2.4,c]X[4,c] X[0.5,c]}
	\toprule
	\multirow{2}{*}{\emph{Dataset}} & \multirow{2}{*}{\emph{Area}} & \multirow{2}{*}{\emph{Platform}} & \multicolumn{5}{c}{\emph{Sensors}} & \multirow{2}{*}{\emph{Ground truth}}  & \multirow{2}{*}{\emph{DOF}} & \multirow{2}{*}{\emph{Accuracy}} & \multirow{2}{*}{\emph{Distance}} & \multirow{2}{*}{\emph{Weather}} & \multirow{2}{*}{\emph{Site}} \\
            & & & Lidar & Cam. & IMU & GNSS & Radar & & & & & & \\
		\midrule

            KAIST~\cite{KAIST2018} & Urban & Car & \checkmark & \checkmark & \checkmark & \checkmark & - & INS & 3 & $<$ \SI{10}{\cm}$^{\ast}$ & \SI{191}{\km} & - & O \\

            Wild-Places~\cite{Knights2023} & Forest & Handheld & \checkmark & \checkmark & \checkmark & - & - & HD map & \textbf{6} & $<$ \SI{30}{\cm} & \SI{33}{\km} & - & O \\

            NCLT~\cite{NCLT2015} & Campus & Segway & \checkmark  & \checkmark & \checkmark & \checkmark & - & INS & \textbf{6} &  $\thickapprox$ \SI{10}{\cm}$^{\ast}$ &\SI{147}{\km} & - & I/O \\
                        
		KITTI~\cite{KITTI2012} & Urban & Car & \checkmark & \checkmark & \checkmark & \checkmark & - & INS & \textbf{6} & $<$ \SI{10}{\cm}$^{\ast}$ & \SI{39}{\km} & - & O \\
            
            nuScenes~\cite{nuScenes2020} & Urban & Car & \checkmark & \checkmark & \checkmark & \checkmark & \checkmark & HD map & \textbf{6} &  $<$ \SI{10}{\cm} &\SI{242}{\km} & \textbf{Rain} & O \\

            Oxford~\cite{Oxford2020} & Urban & Car & \checkmark  & \checkmark & \checkmark & \checkmark & \checkmark & INS & \textbf{6} &  $\thickapprox$ \SI{1}{\cm} & \textbf{\SI{280}{\km}} &  \textbf{Fog, rain, snow} & O \\

            Hilti-Oxford~\cite{Hilti2023} & Urban & Handheld & \checkmark & \checkmark & \checkmark & - & - & HD map & \textbf{6} &  $<$ \SI{1}{\cm} &$<$ \SI{10}{\km} & - & I/O \\
            
            \midrule 

            Hilti SLAM~\cite{Hilti2022} & Urban & Handheld & \checkmark & \checkmark & \checkmark & - & - & Hilti PLT 300 & 3 &  $\thickapprox$ \SI{3}{\mm} &$<$ \SI{10}{\km} & - & I/O \\
            
            Euroc~\cite{Euroc2016} & Indoor & UAV & -  & \checkmark & \checkmark & - & - & Leica MS50 & 3 &  \textbf{$\thickapprox$ \SI{1}{\mm}} & \SI{0.9}{\km} & - & I \\
            
            UZH-FPV~\cite{UZH2019} & Campus & UAV & -  & \checkmark & \checkmark & - & - & Leica MS60 & 3 & \textbf{$\thickapprox$ \SI{1}{\mm}} & \SI{10}{\km} & - & I \\ 
            
            TUM~\cite{TUMVI2018} & Campus & Handheld & - & \checkmark & \checkmark & - & - & OptiTrack & \textbf{6} & \textbf{$\thickapprox$ \SI{1}{\mm}} & $<$ \SI{1}{\km}* & - & I/O \\
            
            Challenging dataset~\cite{Challenge2012} & Campus, mountain & Rig & \checkmark  & - & \checkmark & - & - & Leica TS15 & \textbf{6} & \textbf{$\thickapprox$ \SI{1}{\mm}} & $<$ \SI{1}{\km} & - & I/O \\ 
            
            \emph{\dataset{} dataset (Ours)} & Campus, forest & UGV & \checkmark & - & \checkmark & \checkmark & - & Trimble S7 & \textbf{6} & $\thickapprox$ \SI{4}{\mm} & \textbf{\SI{49}{\km}} & \textbf{Fog, rain, snow} & I/O \\
		\bottomrule
	\end{tabu}
 \vspace{-1mm}
\end{table*}

Among the majority of datasets used for benchmarking \ac{SLAM} algorithms, the prominent reference trajectory generation method is \ac{GNSS}-Aided \ac{INS}.
This approach relies on integrating data from one or more GNSS receivers with high-frequency measurements obtained from a \ac{IMU} within an \ac{INS}~\cite{grewal2007global}.
This fusion provides the six-\ac{DOF} pose estimation for the platform.
The KITTI dataset~\cite{KITTI2012} was the first dataset to introduce a \ac{SLAM} algorithm benchmark using over \SI{39}{\km} of reference trajectory data, generated by an \ac{GNSS}-Aided \ac{INS}. 
This system was mounted on a car equipped with lidar and camera sensors for continuous data collection.
Given this success within the scientific community, numerous other benchmarks emerged, employing sensor-equipped cars and similarly utilizing \ac{GNSS}-Aided \ac{INS} for production of reference trajectories datasets~\cite{KAIST2018,Oxford2020,NCLT2015}. 
Nonetheless, \ac{GNSS}-Aided \ac{INS} systems encounter challenges in urban environments, such as the \ac{GNSS} canyon effect caused by buildings, issues related to signal multi-path, and limited satellite visibility. 
As a result, the system's accuracy can fluctuate, typically ranging around \SI{10}{\cm} which may be too significant for specific benchmarking scenarios such as the TUM-VI dataset~\cite{TUMVI2018}.
This dataset focused on high-speed drone localization using camera images and \ac{IMU} data.
Given the drone's aggressive dynamics, a one-millimeter-accurate Optitrack system was used to generate the six-\ac{DOF} reference trajectory.
Other datasets have employed high-definition 3D maps provided by terrestrial scanners to generate reference trajectories~\cite{nuScenes2020}. 
The obtained accuracy is under ten centimeters, and the dataset supports object detection and tracking to train autonomous vehicle algorithms under clear or rainy conditions. 
More recently,~\cite{Hilti2023} also utilizes high-resolution mapping of urban locations to reconstruct a camera, lidar, and \ac{IMU} setup's trajectory in six-\ac{DOF}. 
Lidar scans are directly aligned with the high-definition 3D map, resulting in millimeter-level accuracy.
For natural environments,~\cite{Knights2023} used an \ac{SLAM} algorithm to reconstruct their setup's trajectory in six-\ac{DOF} and used it as a ground truth with an \SI{30}{\cm} accuracy.
All of these methods are limited in accuracy, as seen with \ac{GNSS}-Aided \ac{INS} systems, or are difficult to apply in outdoor environments, such as high-resolution terrestrial laser scanning or Optitrack measurement methods. 
Our \dataset{} dataset overcomes these limitations, offering a dataset in various indoor and outdoor environments with centimeter-level accuracy through the use of multiple \acp{RTS}.

As mentioned earlier, previous datasets have already employed \acp{RTS} to generate reference trajectories for their robotic platforms. 
The EuRoc~\cite{Euroc2016} and UZH-FPV~\cite{UZH2019} datasets have employed a Leica \ac{RTS}-based system to generate reference trajectories for drones navigating in indoor and outdoor environments. 
Similarly, the Hilti \ac{SLAM} dataset~\cite{Hilti2022} uses a Leica \ac{RTS}-based system for dynamic tracking of a setup equipped with lidar, cameras, and an \ac{IMU}.
To generate reference trajectories using \acp{RTS}, two strategies exist in the literature. 
The first strategy, employed by~\citet{Challenge2012}, involves placing three passive prisms on a robotic platform and manually capturing static position measurements using an \ac{RTS}.
This method provides the platform's six-\ac{DOF} with millimeter-level accuracy. 
However, the number of obtained reference poses is limited due to the time-consuming nature of manual \ac{RTS} measurements.
The second strategy involves dynamically tracking an active prism with an \ac{RTS}.
The positional measurement accuracy is in the order of millimeters, and the entire prism trajectory can be reconstructed. 
However, the generated reference trajectory has only three-\acp{DOF} as a single prism is tracked in real-time.

The~\autoref{tab:related_work} provides an overview of the mentioned datasets along with their respective characteristics.
To the best of our knowledge, there is not any reference dataset that provides a six-\ac{DOF} reference trajectory of a dynamic platform using a multi-\ac{RTS}-based system.
Moreover, there is a lack of precision information in available datasets regarding the precision of the dynamic \ac{RTS} or \ac{GNSS} measurements.
Hence, we present the \dataset{} dataset, the first dataset focusing on creating six-\ac{DOF} reference trajectories using three \acp{RTS} tracking three distinct active prisms.
Additionally, we provide the precision of measurements and the final poses, a unique feature not previously provided in other datasets.

\section{Hardware}
\label{sec:hardware}

\begin{figure}[htbp]
    \centering
    \includegraphics[width=\linewidth]{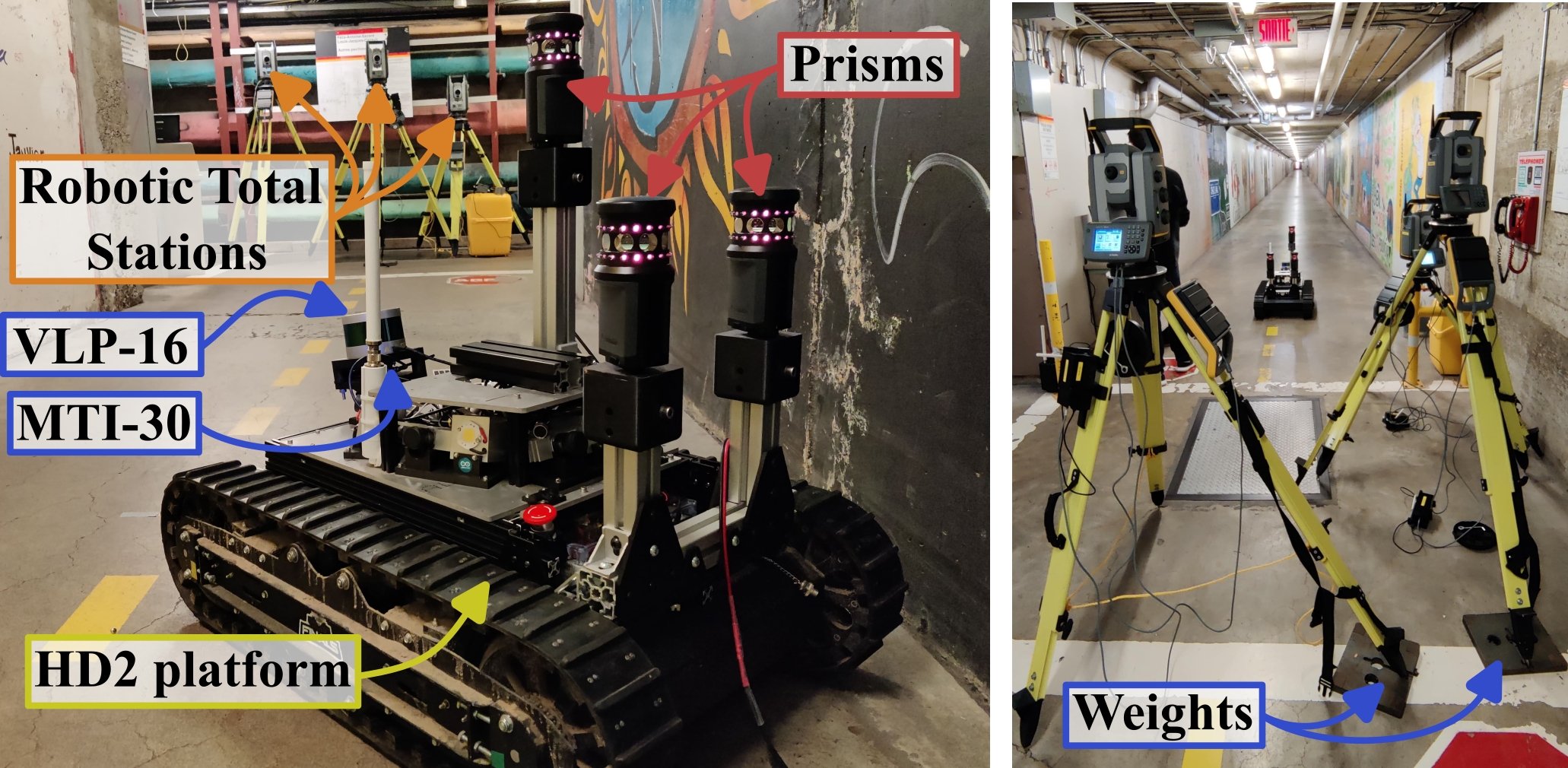}
    \caption{Setup used in the tunnel sites. \emph{Left}: \ac{RTS} setup with the HD2 robot. \emph{Right}: one deployment done in a \SI{120}{\m} tunnel. Because the floor was slippery, heavy weights were added to stabilize the \acp{RTS} tripods.}
    \label{fig:tunnel}
\end{figure}
\begin{figure}[htbp]
    \centering
    \includegraphics[width=\linewidth, trim={0 30 0 37}, clip]{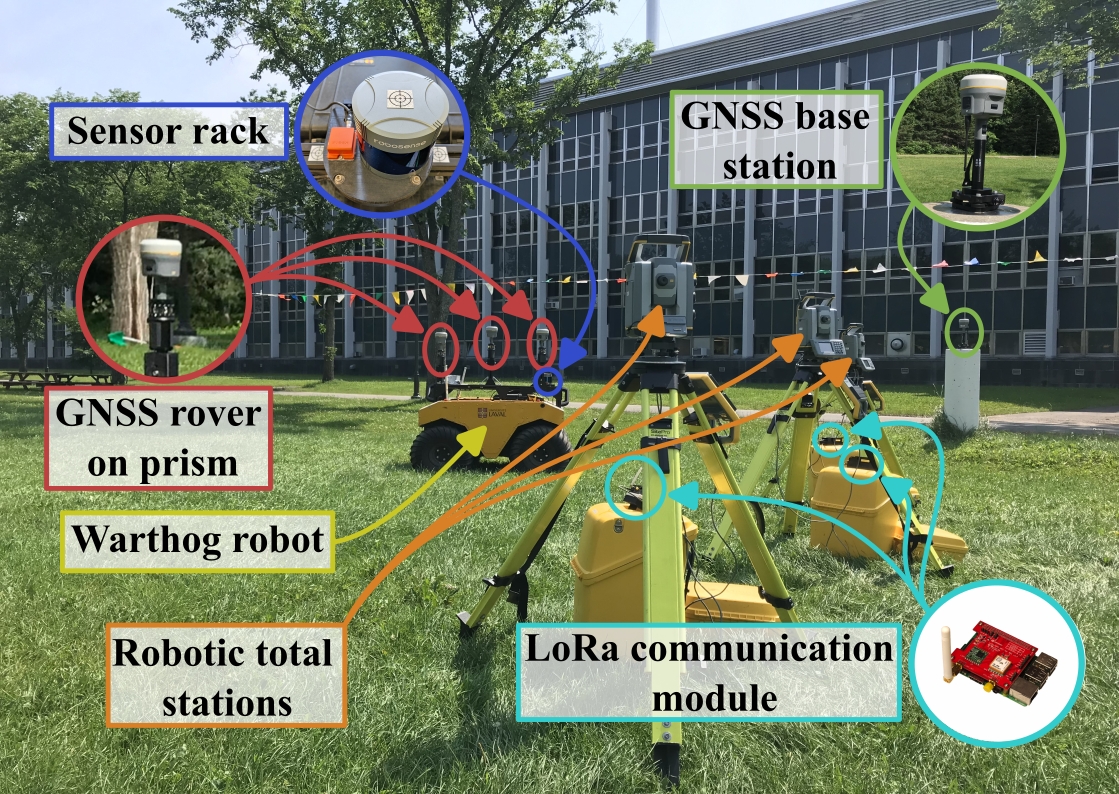}
    \caption{Setup used on the campus with the Warthog \ac{UGV}. 
    A \ac{GNSS} fixed base station sends corrections to the three \ac{GNSS} rovers on the robot. Three prisms are tracked by three \ac{RTS}. 
    Data are collected by three Raspberry Pi clients connected by USB to the \ac{RTS}. 
    A LoRa communication protocol is used to send data to a Raspberry Pi master located on the \ac{UGV}. 
    The lidar and \ac{IMU} are on the front part of the Warthog.}
    \label{fig:harware}
\end{figure}

Two mobile robotic platforms were deployed for the dataset: a Clearpath Warthog \ac{UGV}, and a SuperDroid HD2 \ac{UGV}.
The Warthog robot is equipped with a RoboSense RS-32 lidar and an XSens MTi-10 \ac{IMU}, whereas the HD2 robot's sensor payload consists of a Velodyne-VLP 16 lidar and an XSens MTi-30 \ac{IMU}.
We generalized our approach to generate reference trajectories by using and comparing two different types of setup for the two robots, as shown in~\autoref{fig:tunnel} and~\autoref{fig:harware} for deployments of the HD2 and Warthog platforms, respectively.

The first ground truth setup is composed of three Trimble S7 \acp{RTS}.
Each \ac{RTS} tracks a single Trimble MultiTrack Active Target MT1000 prism at a maximum achievable measurement rate of \SI{2.5}{\Hz}.
For this active prism, the nominal range at the tracking mode is \SI{800}{\m} and the nominal position measurement accuracy is \SI{4}{\mm}.
To collect the data from each \ac{RTS}, a Raspberry Pi 4 is used as a client through a USB connection.
The data are then sent to a master Raspberry Pi 4 through LoRa Shield radio modules from Dragino operating with a radio frequency of \SI{905}{\MHz}.
The chosen modulation allows the Raspberry Pi to reliably send data over a distance of up to \SI{800}{\m} at \SI{366}{\bps} in open space.
The second ground truth setup is composed of a set of four \ac{GNSS} receivers used for outdoor experiences.
In this setup, three \ac{GNSS} receivers are mounted on the Warthog \ac{UGV} and the fourth \ac{GNSS} receiver serves as a static base station. 
For the sake of comparison, two different models of GNSS receivers are used in this dataset, \emph{Reach RS+} and \emph{Trimble R10-2}.
When operating in \ac{RTK} rover/base mode, the \emph{Trimble R10-2} receiver has a vertical accuracy of \SI{21.2}{\mm} and horizontal accuracy of \SI{11.3}{\mm}, while the \emph{Reach RS+} has a vertical accuracy of \SI{19.8}{\mm} and horizontal accuracy of \SI{9.9}{\mm}.
All three prisms and \ac{GNSS} receivers were mounted on the Warthog robot for outdoor experiments. 
The HD2 \ac{UGV} was used to collect data in the university tunnels.
In such environments, \ac{GNSS} receivers are not functional, thus, only the prisms can be used. 

\section{Data collection}
\label{sec:data}


Our main contribution revolves around delivering a dataset featuring openly available ground truth trajectories obtained using diverse setups. 
The \dataset{} dataset was gathered across three distinct environmental settings, encompassing various times of day and spanning around diverse weather conditions, summarized in the~\autoref{tab:experiments}.
\begin{table}[htbp]
        \scriptsize
	\centering
    \vspace{1mm}
	\caption{Table of deployments in the \dataset{} dataset. 
 The weather legend is as follows: C=Clear, FR=Freezing rain, S=Snow, R=Rain. 
 The robot legend means W=Warthog and H=HD2.} 
	\label{tab:experiments}
	\begin{tabu}{X[1,c]X[2,c]X[1,c]X[1,c]X[2,c]X[1,c]X[2,c]}
		\toprule
		  & \emph{Month} & \emph{Exp.} & \emph{Length} & \emph{Weather} & \emph{Robot} & \emph{Setup} \\
		\midrule
	    & Feb. 22 & 1 & \SI{1.58}{\km} & C & W & RTS/GNSS \\
            & Mar. 22 & 6 & \SI{7.45}{\km} & FR, S, C & W & RTS/GNSS \\
            & May 22 & 8 & \SI{14.43}{\km} & C & W & RTS/GNSS \\
            Campus & Jun. 22 & 4 & \SI{3.42}{\km} & C & W & RTS/GNSS \\
            & Jul. 22 & 6 & \SI{2.67}{\km} & C & W & RTS/GNSS \\
            & Nov. 22 & 14 & \SI{3.91}{\km} & C & W & RTS/GNSS \\
            & Dec. 22 & 4 & \SI{2.1}{\km} & C, R & W & RTS/GNSS \\
            & Jul. 23 & 2 & \SI{2}{\km} & C & W & RTS/GNSS \\
            \midrule
            & May 22 & 6 & \SI{1.55}{\km} & C & H & RTS \\
             Tunnel & Jul. 22 & 5 & \SI{1.58}{\km} & C & H & RTS \\
            & Sep. 22 & 9 & \SI{0.85}{\km} & C & H & RTS \\
            \midrule
             Forest & Nov. 22 & 4 & \SI{3.16}{\km} & C & W & RTS/GNSS \\
		\bottomrule
	\end{tabu}
    \vspace{-1mm}
\end{table}

\subsection{Environments}

The first deployment site is the campus of Université Laval in Quebec City. 
This campus comprises buildings, open spaces, and wooded areas.
Multiple views of the campus are depicted in~\autoref{fig:intro}-\emph{right} and~\autoref{fig:harware}.
Overall, \num{45} different experiments through \num{20} distinct deployments were conducted with the Warthog \ac{UGV}, totaling \SI{37.6}{\km} of trajectories on the campus.
During these data collection campaigns, various weather conditions were encountered, including clear weather, rain, and even a snowstorm. 
Data from both \ac{RTS} and \ac{GNSS} setups are available for each of the conducted experiments, enabling a comparison of their respective accuracies in generating reference trajectories.

For the second deployment site, underground tunnels located beneath the university campus were selected.
Four deployments were conducted in these tunnels to gather data from \ac{RTS} setup during \num{20} different experiments, covering a total of \SI{4}{\km} of trajectories with the HD2 platform.
These tunnels have lengths of several hundred meters, which might cause particular challenges for \ac{SLAM} algorithms.
Unlike other setups based on motion capture or ultra-wideband, our \ac{RTS} setup can generate reference trajectories at long ranges without the need to alter the environment.
The~\autoref{fig:tunnel} shows an example of a tunnel as a long-range environment.

Finally, the last site is located in the Montmorency Forest, which belongs to Université Laval. 
This site contains numerous paths for snowmobiles and cross-country ski trails in a dense forest. 
Two deployments were carried out with the Warthog robot, along with \ac{RTS} and \ac{GNSS} systems.
In total, \SI{3.2}{\km} of trajectories were acquired during four different experiments. 
An example of the \ac{RTS} system in the forest is depicted in~\autoref{fig:intro}-\emph{left}.
The lidar was available during the majority of deployments to collect data for \ac{SLAM} algorithms.


\subsection{Ground truth protocol}

In this section, we introduce the standardized protocols that we employed for each of the setups during all conducted deployments, split between \ac{RTS} and \ac{GNSS}.
Starting with \textbf{\ac{RTS} protocol}, the field deployment of the three \acp{RTS} is carried out in several steps:
\begin{enumerate}
    \item (\SI{30}{\min}) \ac{RTS} units are acclimated to the ambient temperature before data collection can begin.
    This step is necessary to prevent any condensation effects that could bias the measurements, especially in winter;
    \item (\num{5}-\SI{60}{\min}) Tripods for the \ac{RTS} are set up, while the \ac{RTS} units adjust to the ambient temperature;
    \item (\num{10}-\SI{15}{\min}) \ac{RTS} units are mounted on tripods, and we roughly level \ac{RTS} units visually.
    To achieve a finer leveling for \ac{RTS} units, calibrated electronic sensors are utilized;
    \item (\num{3}-\SI{5}{\min}) Raspberry Pi devices are powered on and connected via USB to the \ac{RTS} units to retrieve their data and send it to the master unit for recording. 
    These Raspberry Pi devices also assign a unique prism number to each \ac{RTS} unit to be tracked;
    \item (\num{1}-\SI{2}{\min}) Active prisms are mounted on the \ac{UGV} and powered on with their unique ID.
    To facilitate data processing of multiple experiments conducted by the robots, the same prism ID and positions were adopted during all data acquisition operations;
    \item (\num{1}-\SI{10}{\min}) Verification of proper prism tracking is performed, followed by measurements of static prism positions to enable post-processing extrinsic calibration of \ac{RTS} units.
\end{enumerate}
The complete setup process takes between \SI{50}{\min} and \SI{120}{\min}, which depends on the number of available operators and weather conditions. 
At the end of the data collection procedure, prisms have to remain on the \ac{UGV} to immediately perform an extrinsic sensor calibration, discussed later.

As for the \textbf{\ac{GNSS} protocol}, we need a minimum of three \ac{GNSS} receivers as rover plus one \ac{GNSS} receiver as a base station to generate a six-\ac{DOF} ground truth with only \ac{GNSS}-based setup.
This rover/base configuration, known as \ac{RTK}, can achieve centimeter-level accuracy by correcting errors of the GNSS receivers used as rovers.
The base station, consisting of a \ac{GNSS} receiver, provides corrections for the rover \ac{GNSS} observations by simultaneously monitoring the same satellites as the rover receivers.
The base station can be fixed at a predetermined, known, and stationary location (\eg a geodesic pillar stock or a geodetic survey marker) or an unknown position.
If the position of the base station is known in advance, \ac{GNSS} receivers have just to be powered on and once the radio connection between the base and rovers is established the system is operational and the collection of data can start. 
On the other hand, if the position of the base station is unknown, a waiting time of at least \SI{15}{\min} is necessary before data collection to let the rover or the base receivers have enough time to boot, average their positions, and achieve optimal reading of the satellite constellations in the sky.
The positional corrections are then transmitted through messages via a radio link from the base station to the rover receivers, where they are employed to correct the real-time positions of the rover.
Moreover, the internal radio of the \ac{GNSS} receiver has a maximum range of \SI{2}{\km}. 
To achieve a greater range, we have to use the external radio, which can theoretically transmit up to \SI{10}{\km} under optimal conditions.


\subsection{Calibration and time synchronization}

This section specifies how the systems were calibrated and synchronized, along with the data format given by the dataset.
Attention is given to 1) extrinsic calibration and 2) time synchronization.
First, the extrinsic calibration process involves obtaining the precise pose of all the sensors on the robots. 
Calibration is performed using one \ac{RTS} at the end of each deployment to closely match the conditions of the experiments as temperature, pressure, and humidity can affect measurements, especially in winter. 
Retro-reflective targets, as shown in~\autoref{fig:harware}, are stuck atop the lidar and \ac{GNSS} receivers.
The prisms are left in active mode for this calibration.
Ten repetitions are performed for each millimeter-precise position to increase precision and to determine their uncertainties.
This method provides the relative positions of the prisms, \ac{GNSS} receivers, and lidar to each other. 
Since the \ac{IMU} is too small, its extrinsic calibration with the lidar is based on the work of~\citet{Kubelka2022}, who uses lidar and \ac{IMU} data to perform calibration within a tenth of a degree using a modified four-\ac{DOF} \ac{SLAM} algorithm. 
Translation between the lidar and \ac{IMU} is determined using the \ac{CAD} model of their support.
These calibrations were performed after each deployment to ensure precise referenced measurements.
As for time synchronization, a \ac{NTP} daemon was used to synchronize the clocks between the two on-board computers on the robot (\ie the main computer of the \ac{UGV} and the Raspberry Pi master connected to the robot network).
Ten minutes was allowed after booting up both of the computers so that the client clock could adjust.
Measurements from the wheel encoders, lidar, and \ac{IMU} were timestamped using the data logging computer’s clock.
Synchronization between the master and client Raspberry Pi is achieved using a modified \ac{NTP} protocol designed for LoRa communication, as described in~\cite{Vaidis2021}. 
Initial synchronization is performed at the start of data collection, followed by repetitions every five minutes for each client. 
This method ensures clock precision at the level of one to two milliseconds.
As the \ac{GNSS} devices are not connected to the setup, temporal synchronization to the \ac{RTS} data is established using a classical maximum likelihood state estimator, similar to the approach taken by~\citet{Euroc2016}.


\subsection{Data format}

For each deployment and experiment, ROS~2 rosbag files are provided for data gathered by the Warthog and HD2 \ac{UGV}.
These files include lidar scans, \ac{IMU} measurements, motors and encoders data, as well as rigid transformation between all sensor frames.
\ac{URDF} files are provided along with these rosbag files for each \ac{UGV}.
Note that no \ac{INS} were used to process the \ac{IMU} measurements.
All \ac{IMU} raw data is available in the different rosbag files.
With the \ac{RTS} setup, prism positions, and time synchronizations are provided for rosbag files in ROS~1 and ROS~2. 
The static prism positions computed for the \ac{RTS} extrinsic calibration are given in a text file \texttt{GCP.txt}, and the sensor extrinsic calibration results are given in another text file \texttt{calibration\_raw.txt}.
Finally, \texttt{NMEA} and \texttt{UBX} files are provided for all \ac{GNSS} receivers used during experiments.
The \dataset{} dataset is available at \url{https://github.com/norlab-ulaval/RTS_project}, along with a toolbox code to process the data.

\section{Discussion and challenges}
\label{sec:results}

To assess the disparities among the setups, we conducted an analysis of \num{27} outdoor experiments, covering a total distance of \SI{20.6}{\kilo\meter}, during which we tracked the trajectories of the robot using \ac{GNSS} and \ac{RTS}-based setups when available.
We evaluated the precision of each system by employing \emph{inter-distances} between prisms and \acp{GNSS}.
These distances are calculated between every synchronized triplet of prism positions or \ac{GNSS} receiver positions recorded during each experiment, respectively referred as inter-prism distances and inter-\ac{GNSS} distances.
Subsequently, each of these distance triplets is compared to their corresponding calibrated distance (evaluated at step six of our protocol) and taken as a reference to obtain the errors in precision.
Furthermore, we employed an inter-precision distance to quantify variation in precision between different experiments conducted on the same site at different times.
The purpose of this distance is to highlight the reproducibility of the different setups to generate ground truth trajectories. 
Sets of closest positions in the ground truth trajectories recorded during separate experiments are computed by a nearest neighbors algorithm. 
Subsequently, the inter-precision distance is computed for each set by subtracting each average inter-prism or inter-\ac{GNSS} distances of each position.

\subsection{Precision and reproducibility}

Results of inter-prism distances, inter-\ac{GNSS} distances, and precision on the final translation and rotation of the robot are shown in~\autoref{tab:precision} for \num{15} deployments.
The precision on the final pose is estimated by the same method used in our previous work~\cite{Vaidis2023Iros}.
It can be seen that \ac{RTS} precision is stable in multiple environments, while the \ac{GNSS} can have a variation of more than \SI{300}{mm} in the forest environment, as well as in open space.
Moreover, higher distances between a prism and its \ac{RTS} affect the precision, leading to higher uncertainties. 

\renewcommand{\opacity}{100}
\def\angle{0}
\tabulinesep=0.2mm
\begin{table*}[!htbp] 
    \centering
    \vspace{1mm}
    \caption{Median values for different metrics obtained over \num{15} \ac{RTS} and \ac{GNSS} setups.
    All deployments happened during 2022 and the date format is day/month.
    For each row, light colors indicate low values, whereas dark means high values.}
    \label{tab:precision}
    \centering
    \begin{tblr}{
        colspec = {X c*{14}{p{5mm}}},
        vline{11,13} = {solid},
    }
    & \SetCell[c=15]{c} \emph{Environment}
    \\
    \cline[1pt]{2-Z}
    &
    \SetCell[c=9]{c} \emph{Open space} &&&&&&&&&
    \SetCell[c=2]{c} \emph{Building} &&
    \SetCell[c=4]{c}\emph{Forest} &&&
    \\ 
     &
    \rotatebox{\angle}{24/02} &
    \rotatebox{\angle}{07/03} &
    \rotatebox{\angle}{14/03} &
    \rotatebox{\angle}{16/03} &
    \rotatebox{\angle}{22/06} &
    \rotatebox{\angle}{30/06} &
    \rotatebox{\angle}{11/07} &
    \rotatebox{\angle}{29/11} &
    \rotatebox{\angle}{05/12} &
    \rotatebox{\angle}{12/03} &
    \rotatebox{\angle}{16/11} &
    \rotatebox{\angle}{31/03} &
    \rotatebox{\angle}{09/11} &
    \rotatebox{\angle}{10/11} &
    \rotatebox{\angle}{24/11} 
    \\ 
    \hline[0.5pt]
    Inter-prism distance [mm] & \SetCell{bg=red7, fg=black}6.5 & \SetCell{bg=red9, fg=black}2.4 & \SetCell{bg=red8, fg=black}4.1 & \SetCell{bg=red8, fg=black}4.1 & \SetCell{bg=red6, fg=black}9.7 & \SetCell{bg=red8, fg=black}5.7 & \SetCell{bg=red1, fg=white}22.9 & \SetCell{bg=red8, fg=black}4.6 & \SetCell{bg=red8, fg=black}3.7 & \SetCell{bg=red8, fg=black}4.3 & \SetCell{bg=red8, fg=black}4.5 & \SetCell{bg=red7, fg=black}6.6 & \SetCell{bg=red3, fg=white}18.0 & \SetCell{bg=red8, fg=black}5.0 & \SetCell{bg=red9, fg=black}3.2 \\
    Inter-GNSS distance [mm] & \SetCell{bg=red9, fg=black}5.5 & \SetCell{bg=red9, fg=black}5.2 & \SetCell{bg=red9, fg=black}4.6 & \SetCell{bg=red9, fg=black}6.4 & \SetCell{bg=red8, fg=black}37.2 & \SetCell{bg=red2, fg=white}368.0 & \SetCell{bg=red7, fg=black}111.0 & \SetCell{bg=red9, fg=black}6.9 & \SetCell{bg=red9, fg=black}20.5 & \SetCell{bg=red5, fg=black}230.0 & \SetCell{bg=red9, fg=black}17.7 & \SetCell{bg=red4, fg=white}288.0 & \SetCell{bg=red6, fg=black}149.0 & \SetCell{bg=red2, fg=white}394.0 & \SetCell{bg=red1, fg=white}423.0 \\
    Translation error [mm] & \SetCell{bg=red7, fg=black}15.2 & \SetCell{bg=red3, fg=white}28.7 & \SetCell{bg=red4, fg=white}25.3 & \SetCell{bg=red8, fg=black}10.6 & \SetCell{bg=red7, fg=black}12.3 & \SetCell{bg=red7, fg=black}12.8 & \SetCell{bg=red1, fg=white}35.2 & \SetCell{bg=red4, fg=white}22.9 & \SetCell{bg=red7, fg=black}12.7 & \SetCell{bg=red4, fg=white}23.7 & \SetCell{bg=red8, fg=black}12.0 & \SetCell{bg=red6, fg=black}19.0 & \SetCell{bg=red7, fg=black}12.8 & \SetCell{bg=red9, fg=black}6.7 & \SetCell{bg=red5, fg=black}21.0 \\
    Rotation error [deg] & \SetCell{bg=red7, fg=black}1.51 & \SetCell{bg=red7, fg=black}1.6 & \SetCell{bg=red5, fg=black}2.05 & \SetCell{bg=red7, fg=black}1.4 & \SetCell{bg=red8, fg=black}0.9 & \SetCell{bg=red8, fg=black}0.94 & \SetCell{bg=red1, fg=white}4.0 & \SetCell{bg=red6, fg=black}1.94 & \SetCell{bg=red7, fg=black}1.33 & \SetCell{bg=red6, fg=black}1.93 & \SetCell{bg=red8, fg=black}0.85 & \SetCell{bg=red8, fg=black}1.06 & \SetCell{bg=red8, fg=black}0.97 & \SetCell{bg=red9, fg=black}0.53 & \SetCell{bg=red7, fg=black}1.59 \\
    \ac{RTS} Range [m] & \SetCell{bg=red7, fg=black}39.0 & \SetCell{bg=red9, fg=black}20.0 & \SetCell{bg=red9, fg=black}18.0 & \SetCell{bg=red9, fg=black}19.0 & \SetCell{bg=red1, fg=white}90.0 & \SetCell{bg=red7, fg=black}32.0 & \SetCell{bg=red1, fg=white}87.0 & \SetCell{bg=red7, fg=black}39.0 & \SetCell{bg=red7, fg=black}32.0 & \SetCell{bg=red3, fg=white}71.0 & \SetCell{bg=red6, fg=black}42.0 & \SetCell{bg=red7, fg=black}38.0 & \SetCell{bg=red6, fg=black}42.0 & \SetCell{bg=red4, fg=white}59.0 & \SetCell{bg=red7, fg=black}37.4 \\
    \hline
    \end{tblr}
    \vspace{-1mm}
\end{table*}

\begin{figure}[htbp]
    \centering
    \includegraphics[width=\linewidth]{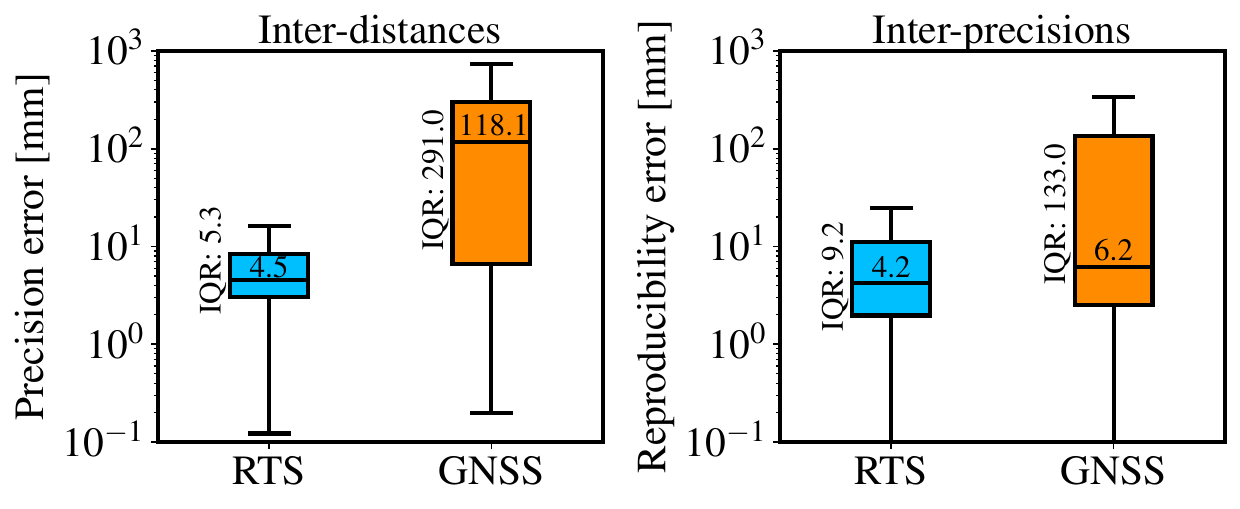}
    \caption{
    Distribution of errors arising for the two setups. \emph{Left}: inter-prism and inter-\ac{GNSS} distances. \emph{Right}: inter-precision distances.
    Results obtained from the \acp{RTS} are denoted in blue, whereas those from the \ac{GNSS} are shown in orange.
    The median error values are in the center of each box, and the \ac{IQR} is indicated alongside for reference.}
    \label{fig:results}
\end{figure}

The results depicted in~\autoref{fig:results}-\emph{Left} offer general insights into the errors associated with the inter-prism and inter-\ac{GNSS} distances. 
These findings reveal that the \ac{RTS} acquisition system consistently achieves a median precision of approximately \SI{4.5}{\milli\meter}, whereas the \ac{GNSS} system exhibits a \num{22} times lower median precision, being around \SI{118.1}{\mm}.
It is important to emphasize that the inter-distance errors highlight the highest precision of the \ac{RTS} acquisition system compared to the \ac{GNSS} system.
This discrepancy can be attributed to the relatively low error inherent to the \ac{RTS} acquisition system, in contrast to the absolute error associated with \ac{GNSS}.

To evaluate the reproducibility between experiments, we employ a nearest neighbor distance calculation within three meters, with data expressed in the \ac{GNSS} frame. 
As illustrated in~\autoref{fig:results}-\emph{Right}, the \ac{RTS} setup consistently exhibits high reproducibility, with a median error of \SI{4.2}{\milli\meter}.
The \ac{GNSS} have the same level of reproducibility, with a median of \SI{6.2}{\milli\meter}.
However, the \ac{GNSS} \ac{IQR}, with a value of \SI{133}{\milli\meter}, is \num{14} times more important than the \ac{RTS}.
This observation underscores the stability of precision across all experiments for the \ac{RTS} compared to the \ac{GNSS}.

\subsection{Challenges encountered}

\begin{figure}[htbp]
    \centering
    \includegraphics[width=\linewidth, trim={0 20 0 20}, clip]{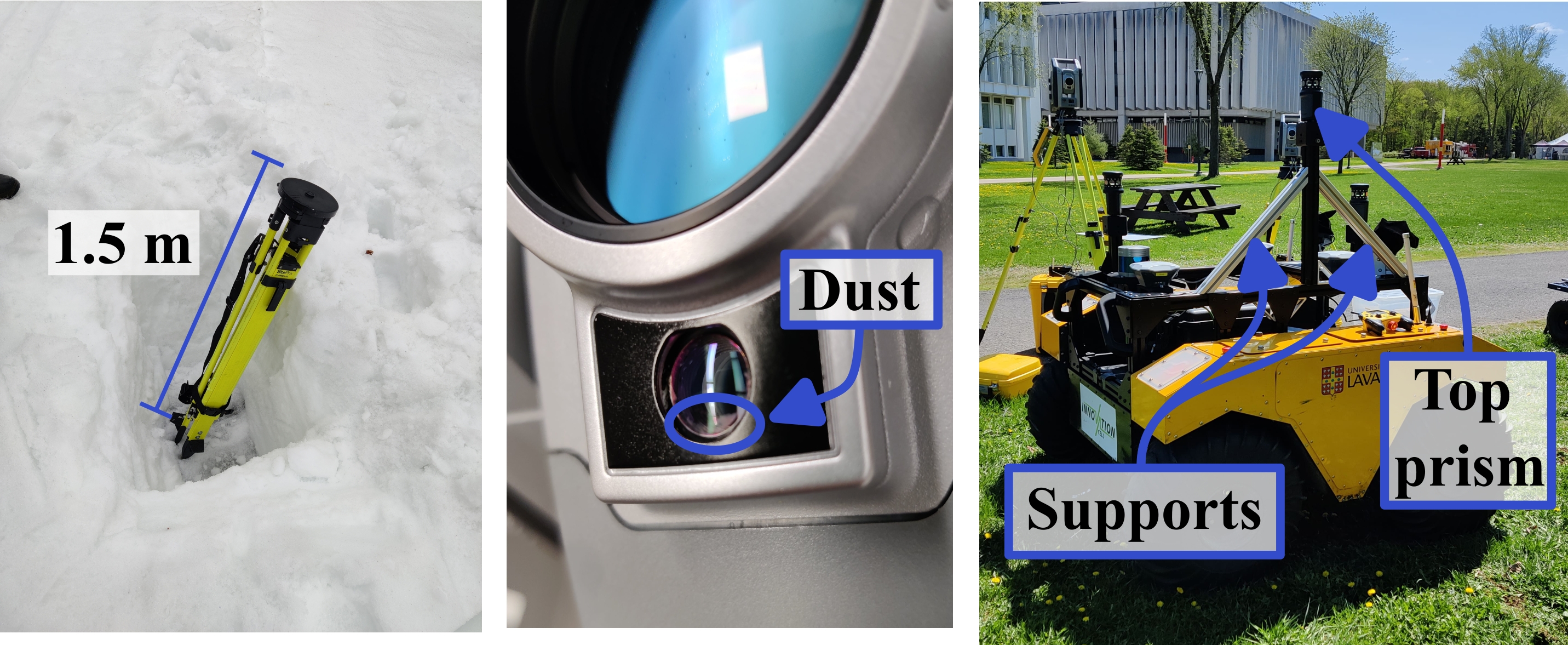}
    \caption{Issues encountered while collecting the dataset. \emph{Left}: difficulties of leveling on deep snow. 
    \emph{Middle}: dust on a lens which was interfering with the tracking mode of the \ac{RTS}. \emph{Right}: supports added on the Warthog to reduce vibrations on the top prism.}
    \label{fig:issues}
\end{figure}

The first encountered challenge was the leveling of the \acp{RTS} in winter.
It is sometimes necessary to remove snow from the ground for step two of our protocol to prevent tripods from gradually sinking into the snow, which would impact the ground truth.
In forested areas where the snow depth can reach several meters, as shown in~\autoref{fig:issues}-\emph{Left}, surface snow is compacted to provide stability.
Secondly, as depicted in~\autoref{fig:intro}, obstacles between prisms and \acp{RTS} can disrupt measurements. 
To address this issue, \ac{RTS} placement locations are pre-selected based on the planned trajectory of the \ac{UGV}, and the heights of both \acp{RTS} and prisms are varied to minimize occlusion risks.
Another challenge is the presence of dust or dirt on \ac{RTS} lenses, as shown in~\autoref{fig:issues}-\emph{Middle}, which can degrade \ac{RTS} performance.
Therefore, lenses should be wiped regularly.
Furthermore, since prisms and \ac{GNSS} units are elevated on the \ac{UGV}, they are susceptible to vibrations during motions. 
These vibrations can lead to positioning errors of up to \SI{1}{\cm}. 
To deal with this problem, metal supports have been added to dampen vibrations, as depicted in~\autoref{fig:issues}-\emph{Right}.
Finally, unlike \ac{GNSS} data, \ac{RTS} measurement timestamps are not globally valid.
To address this issue and obtain the six-\ac{DOF} pose, data interpolation is performed. 
This interpolation can reduce the accuracy of the final estimated pose, especially for high \ac{UGV} dynamic and low-rate measurement setups.

\section{Conclusion}
\label{sec:conclusion}

In this paper, we have introduced a novel dataset, the \dataset{} dataset, designed to compare various ground truthing setups. 
The \dataset{} dataset stands out as a unique dataset for providing six-\ac{DOF} reference trajectories of a moving robotic platform generated by three \acp{RTS}.
The dataset encompasses \ac{GNSS} and \ac{RTS} data as well, along with information from lidar, \ac{IMU}, and encoder sensors.
It facilitates the application of \ac{SLAM} algorithms and the assessment of results to the ground truth data.
Furthermore, the \dataset{} dataset covers diverse environments and weather conditions to assess the quality of the different ground truth setups.
Additionally, tools for quantifying their precision are provided, a feature not present in previous datasets.
An extensive analysis of the precision and reproducibility of different trajectory generation setups was conducted. 
The results indicate that \ac{RTS} systems deliver more precise and reproducible data compared to \ac{GNSS} solutions, even when used in \ac{RTK} mode.
These results demonstrate the challenges posed by generating six-\ac{DOF} ground truth trajectories in outdoor and indoor environments.
Moreover, it shows that multiple \acp{RTS} can be used to benchmark six-\ac{DOF} \ac{SLAM} algorithm comparisons.





\IEEEtriggeratref{6}
\IEEEtriggercmd{\enlargethispage{-0.1in}}

\printbibliography

\end{document}